%% file: main.tex
\documentclass[runningheads]{llncs}
\usepackage{graphicx}
\usepackage{caption}
\usepackage[section]{placeins}
\usepackage{url}
\usepackage[square,sort,comma,numbers]{natbib}

\makeatletter
\AtBeginDocument{%
  \def\doi#1{\url{#1}}}
\makeatother

\begin{document}
\title{
Phylogeny-informed fitness estimation
}

\author{
Alexander Lalejini\inst{1} \and
Matthew Andres Moreno\inst{2} \and
Jose Guadalupe Hernandez \inst{3} \and 
Emily Dolson \inst{3}
}
\authorrunning{A. Lalejini et al.}

\institute{
Grand Valley State University
\and
University of Michigan
\and 
Michigan State University
}

\maketitle              

\input{tex/abstract}

\input{tex/introduction}

\input{tex/phylo-informed-fitness-estimation}

\input{tex/methods}

\input{tex/results-and-discussion}

\input{tex/conclusion}

\input{tex/acknowledgements}

\small
\bibliographystyle{splncs04}
\bibliography{references,supplemental}

\end{document}

%% file: tex/abstract.tex
\begin{abstract}

Phylogenies (ancestry trees) depict the evolutionary history of an evolving population.
In evolutionary computing, a phylogeny can reveal how an evolutionary algorithm steers a population through a search space, illuminating the step-by-step process by which any solutions evolve. 
Thus far, phylogenetic analyses have primarily been applied as post-hoc analyses used to deepen our understanding of existing evolutionary algorithms.
Here, we investigate whether phylogenetic analyses can be used at runtime to augment parent selection procedures during an evolutionary search. 
Specifically, we propose phylogeny-informed fitness estimation, which exploits a population's phylogeny to estimate fitness evaluations. 
We evaluate phylogeny-informed fitness estimation in the context of the down-sampled lexicase and cohort lexicase selection algorithms on two diagnostic analyses and four genetic programming (GP) problems.
Our results indicate that phylogeny-informed fitness estimation can mitigate the drawbacks of down-sampled lexicase, improving diversity maintenance and search space exploration. 
However, the extent to which phylogeny-informed fitness estimation improves problem-solving success for GP varies by problem, subsampling method, and subsampling level. 
This work serves as an initial step toward improving evolutionary algorithms by exploiting runtime phylogenetic analysis.  

\keywords{
genetic programming  \and 
parent selection \and 
phylogeny \and 
lexicase selection.
}

\end{abstract}

%% file: tex/introduction.tex
\section{Introduction}
\label{sec:introduction}

A phylogeny (ancestry tree) details the evolutionary history of a population.
Phylogenetic trees represent the evolutionary relationships among taxa (e.g., individuals, genotypes, genes, species, \textit{etc.}).
In evolutionary biology, phylogenies are typically estimated from the fossil record, phenotypic traits, and extant genetic information. 
Such imperfect phylogenies have profoundly advanced our understanding of the evolution of life on Earth, helping us to organize our knowledge of biological diversity and providing insights into population-level evolutionary dynamics or events that occurred during evolution~\cite{tucker_guide_2017,rothan_epidemiology_2020}.

In evolutionary computing, we can often track phylogenies at runtime with perfect (or near-perfect) accuracy~\cite{moreno_hstrat_2022,ofria_empirical_2020,de_rainville_deap_2012,bohm_mabe_2017,crepinsek_analysis_2011}.
In this context, a phylogeny can reveal how an evolutionary algorithm steers a population through a search space, showing the step-by-step process by which any solutions evolved. 
Indeed, phylogenetic analyses have yielded insights for evolutionary computing research~\cite{mcphee_analysis_1999,donatucci_analysis_2014,burlacu_population_2023,mcphee_visualizing_2017}.
For example, recent work suggests that phylodiversity metrics provide an improved window into the mechanisms that cause evolutionary algorithms to succeed or fail  ~\cite{dolson_ecological_2018,hernandez_phylogenetic_metrics_2022}.
A lack of diversity in a population can cause an evolutionary algorithm to prematurely converge to sub-optimal solutions from which it cannot escape~\cite{goldberg_genetic_1987}. 
Most commonly, an evolutionary algorithm's capacity for diversity maintenance is measured by counting the number of distinct candidate solutions in the population at individual single time points (e.g., counting the number of unique genotypes or phenotypes). 
Phylodiversity metrics, however, take into account the evolutionary history of a population by quantifying the topology of its phylogeny~\cite{hernandez_phylogenetic_metrics_2022,dolson_interpreting_2020}. 
In this way, phylodiversity metrics better captures how well an evolutionary algorithm has explored a search space as compared to other, more commonly used diversity metrics.
The power of this approach is evidenced by the fact that phylodiversity has been shown to be predictive of an evolutionary algorithm's success; that is, in many contexts, runs of evolutionary computation that maintain more phylogenetic diversity are more likely to produce a high-quality solution~\cite{hernandez_phylogenetic_metrics_2022, shahbandegan_untangling_2022}.

Phylogeny-based analyses are primarily applied post-hoc to deepen our understanding of existing evolutionary algorithms. 
Here, we investigate whether phylogenetic analyses can be used at runtime to augment parent selection procedures during an evolutionary search. 
Specifically, we augment down-sampled lexicase selection~\cite{hernandez_random_2019} by exploiting a population's phylogeny to estimate fitness.

Lexicase-based parent selection algorithms have been shown to be highly successful for finding effective solutions to test-based problems across many domains~\cite{la_cava_epsilon-lexicase_2016,moore_lexicase_2017,aenugu_lexicase_2019,metevier_lexicase_2019,lalejini_artificial_2022,ding_optimizing_2022,matsumoto_faster_2023}.
Test-based problems use a large set of input/output examples that specify expected behavior (training cases) to assess the quality of a candidate solution.
Many traditional selection procedures aggregate performance across this training set to produce a single fitness score to be used to select parents.
Lexicase selection avoids aggregation and considers each training case separately, which has been shown to improve diversity maintenance~\cite{helmuth_effects_2016,dolson_ecological_2018} and overall search space exploration~\cite{hernandez_exploration_2022,hernandez_suite_2022}. 
Standard lexicase selection, however, has the drawback of requiring each candidate solution in a population to be evaluated against all training cases, which can be computationally expensive. 
Down-sampled lexicase selection and cohort lexicase selection address this drawback by randomly subsampling the set of training cases used to evaluate candidate solutions each generation, reducing the number of per-generation training case evaluations required.
As a result, more computational resources can be allocated to other aspects of the evolutionary search (e.g., running for more generations), which has been shown to dramatically improve problem-solving success on many problems \cite{hernandez_random_2019,hernandez_characterizing_2020,helmuth_problem-solving_2022,helmuth_benchmarking_2020}. 

Random subsampling can leave out important training cases, causing lexicase selection to fail to maintain important genetic diversity~\cite{hernandez_exploration_2022,boldi_static_2023}.
This drawback is often mitigated by using moderate subsample rates or using a training set with many redundant training cases (e.g., program synthesis).
However, repeated loss of important diversity caused by random subsampling can prevent problem-solving success on problems that require substantial search space exploration or when an expected behavior is represented by a small number of training cases in the training set (making them more likely to be left out in a random subsample).
Recently, Boldi et al. \cite{boldi_informed_2023} proposed \textit{informed} down-sampled lexicase selection, which uses runtime population statistics to construct subsamples that are less likely to leave out important training cases.
Informed down-sampled lexicase improved problem-solving success over naive down-sampling for some problems but did not entirely eliminate the drawbacks of random subsampling~\cite{boldi_static_2023}. 

Here, we propose phylogeny-informed fitness estimation as an alternative approach to mitigating the negative effects of random subsampling.
Phylogeny-informed fitness estimation requires the population's ancestry information to be tracked during evolution.
Each generation, random subsampling is applied to the training set, determining which training cases each candidate solution is directly evaluated against. 
The population's phylogeny is then used to estimate a candidate solution's performance on any training cases that they were not evaluated against, receiving the score of the nearest ancestor (or relative) that \textit{was} evaluated on that training case.
All training cases can then be used for lexicase selection, as each candidate solution has a score for all training cases (either true or estimated).
Of course, phylogeny-informed fitness estimation requires that an individual's score on a particular training case is likely to be similar to that of its close relatives (e.g., parent or grandparent). 
Under low to moderate mutation rates, we hypothesize that this condition holds for many common representations used across evolutionary computing.

In this work, we demonstrate two methods of phylogeny-informed estimation, ancestor-based estimation and relative-based estimation, in the context of both down-sampled lexicase selection and cohort lexicase selection.  
We used the contradictory objectives and multi-path exploration diagnostics from the DOSSIER suite~\cite{hernandez_suite_2022} to investigate whether either method of phylogeny-informed fitness estimation can mitigate the drawbacks of subsampling in the context of lexicase selection. 
These two diagnostics test a selection procedure's capacity for diversity maintenance and search space exploration, two problem-solving characteristics that subsampling often degrades~\cite{hernandez_exploration_2022,boldi_static_2023}.
Next, we assessed phylogeny-informed estimation on four GP problems from the general program synthesis benchmark suites~\cite{helmuth_general_2015,helmuth_psb2_2021}.  

Overall, we find evidence that our phylogeny-informed estimation methods can help mitigate the drawbacks of random subsampling in the context of lexicase selection. 
Using selection scheme diagnostics, we find that phylogeny-informed estimation can improve diversity maintenance and search space exploration. 
In the context of GP, our results indicate that phylogeny-informed estimation's effect on problem-solving success varies by problem, subsampling method, and subsampling level.

%% file: tex/phylo-informed-fitness-estimation.tex
\section{Phylogeny-informed fitness estimation}
\label{sec:phylogeny-informed-fitness-estimation}

Phylogeny-informed fitness estimation is designed for use on test-based problems in combination with any subsampling routine where each candidate solution is evaluated against a subset of the training set each generation. 
Additionally, this fitness estimation technique assumes access to the current population's phylogeny during runtime. 
For simplicity, we assume an asexually reproducing population (i.e., no recombination) in our description and in our examples; however, phylogeny-informed fitness estimation can be extended to systems with recombination. 

\input{tex/figures/fig-estimation-example}

Each generation, each individual in the population is evaluated against a subset of training cases (as determined by the particular subsampling routine). 
For example, 10\% random down-sampling selects 10\% of the training set (at random) to use for evaluation each generation.
In this example scenario, each individual in the population is evaluated against 10\% of the full training set. 
Without any form of fitness estimation, parent selection would be limited to using the performance scores on the sampled training cases.
With phylogeny-informed fitness estimation, however, the complete training set can be used during selection, as the population's phylogeny is used to estimate a candidate solution's performance on any training case that individual was not directly evaluated against.

We propose two methods of phylogeny-based fitness estimation: \textbf{ancestor-based estimation} and \textbf{relative-based estimation}. 
Ancestor-based estimation limits fitness estimation to using only the ancestors along the focal individual's line of descent. 
To estimate an individual's score on a training case, ancestor-based estimation iterates over the individual's ancestors along its lineage (from most to least recent) until finding the nearest ancestor that \textit{was} evaluated against the focal training case. 
This ancestor's score on the training case is then used as an estimate for the focal individual's score on that training case. 
Figure~\ref{fig:estimation-example} depicts a simple example of ancestor-based estimation. 

Relative-based estimation is similar to ancestor-based estimation, except estimations are not limited to using direct ancestors. 
Instead, we use a breadth-first search starting from the focal individual to find the nearest \textit{relative} in the phylogeny that was evaluated against the focal training case.
Unlike ancestor-based estimation, the source of the estimate can be off of the focal individual's line of descent (e.g., a ``cousin''). 
Relative-based estimation may benefit subsampling procedures that use the full training set partitioned across subsets of the population (e.g., cohort partitioning).
In general, however, ancestor-based estimation can be more effectively optimized, making it more computationally efficient.

Simple ancestor-based estimation can forgo full phylogeny tracking, and instead pass the requisite evaluation information from parent to offspring, which is similar to fitness inheritance approximation techniques for evolutionary multi-objective optimization~\cite{chen_fitness_2002,bui_fitness_2005}.
Our approach to ancestor-based estimation is distinct from fitness inheritance techniques because we evaluate all individuals on a subsample of training cases each generation. 
Other fitness inheritance methods assign a proportion of individuals to inherit the average fitness of their parents, forgoing the need to evaluate those individuals.
Such fitness inheritance techniques proved to be difficult to tune the best inheritance proportion for a given problem~\cite{santana-quintero_review_2010}.

To bound the computational cost of searching the phylogeny, the allowed search distance can be limited, and if no suitable ancestor or relative is found within the search limit, the estimate fails.
Even without bounding, both ancestor-based and relative-based estimation can fail to find a suitable taxon in the phylogeny where the focal training case has been evaluated. 
For example, in the first generation, individuals in the population have no ancestors that can be used for estimates. 
In this work, we assume maximally poor performance (e.g., failure, maximum error, \textit{etc.}) for failed estimates. 
We recognize, however, that future work should explore alternative approaches to handling estimation failures. 
For example, it may be better to favor individuals with unknown scores on a training case over individuals known to fail that test case; that is, known failure should be considered to be worse than unknown performance. 
Similarly, future extensions should also consider decreasing the performance score (or increasing the error) on estimations based on how far away they are found in the phylogeny, as estimations from more distant relatives may be more likely to be inaccurate. 

\subsection{Phylogeny tracking}

Phylogeny-informed estimation methods require (1) runtime phylogeny tracking and (2) that the phylogeny is annotated with the training cases each taxon (i.e., node in the tree) has been evaluated against as well as the results of those evaluations. 
Phylogenies can be tracked at any taxonomic level of organization, be it individual organisms (i.e., individual candidate solutions), genotypes, phenotypes, individual genes, \textit{et cetera}.
To refer generally to the items in a phylogeny, we use the term ``taxa'' (or ``taxon'' for a singular item in a phylogeny). 
In our experiments, taxa represented genotypes, and as such, each taxon might represent multiple individual candidate solutions present in the population or that existed at some point during the run. 

\textit{Complete} phylogeny tracking requires that the ancestral relationships of all taxa be stored for the entire duration of the run. 
This mode of tracking, of course, can become computationally intractable for long runs with large population sizes, as the complete phylogeny can rapidly grow to be too large. 
Fortunately, phylogeny-informed estimation does not require complete phylogenies. 
Instead, we recommend pruning dead (extinct) branches for substantial memory and time savings.
Pruning does not affect ancestor-based estimation but may affect relative-based estimation, as close relatives with no descendants in the extant population will be removed as potential sources for fitness estimates. 
However, we do not expect any potential gains in estimation accuracy from using complete phylogenies for relative-based estimation to outweigh the computational costs of complete phylogeny tracking. 

Phylogenies can be tracked without increasing the overall time complexity of an evolutionary algorithm; adding a new node to the tree is a constant-time operation. 
Without pruning, however, the space complexity of phylogeny tracking is $\mathcal{O}$(generations * population size). 
In most cases, pruning reduces the space complexity to $\mathcal{O}$(generations + population size), although the exact impact depends on the strengths of both selection and any diversity maintenance mechanisms being used. 
While worst-case time complexity of pruning is $\mathcal{O}$(generations), this worst-case can only occur a constant number of times per run of an evolutionary algorithm, and in practice pruning is usually fairly fast. 
Thus, while phylogeny tracking does incur a slight computational cost, it can be done fairly efficiently.
Practitioners must determine if the potential benefits of phylogeny tracking (e.g., phylogeny-informed estimation) outweigh the costs for their particular system. 
We expect the potential benefits to outweigh the costs of tracking in systems where fitness evaluation is expensive.
 
Efficient software libraries with tools for phylogeny tracking in evolutionary computing systems are increasingly being developed and made available, such as hstrat~\cite{moreno_hstrat_2022}, 
Phylotrackpy~\cite{dolson_phylotrackpy_2023},
MABE~\cite{bohm_mabe_2017},
Empirical~\cite{ofria_empirical_2020}, 
and
DEAP~\cite{de_rainville_deap_2012}.
In this work, we use the phylogeny tracking from the Empirical C++ library. 
The C++ implementation of our phylogeny-informed estimation methods are available as part of our supplemental material~\cite{supplemental_gh}.

%% file: tex/figures/fig-estimation-example.tex
\begin{figure*}[hbt!]
\begin{center}
\includegraphics[width=\textwidth]{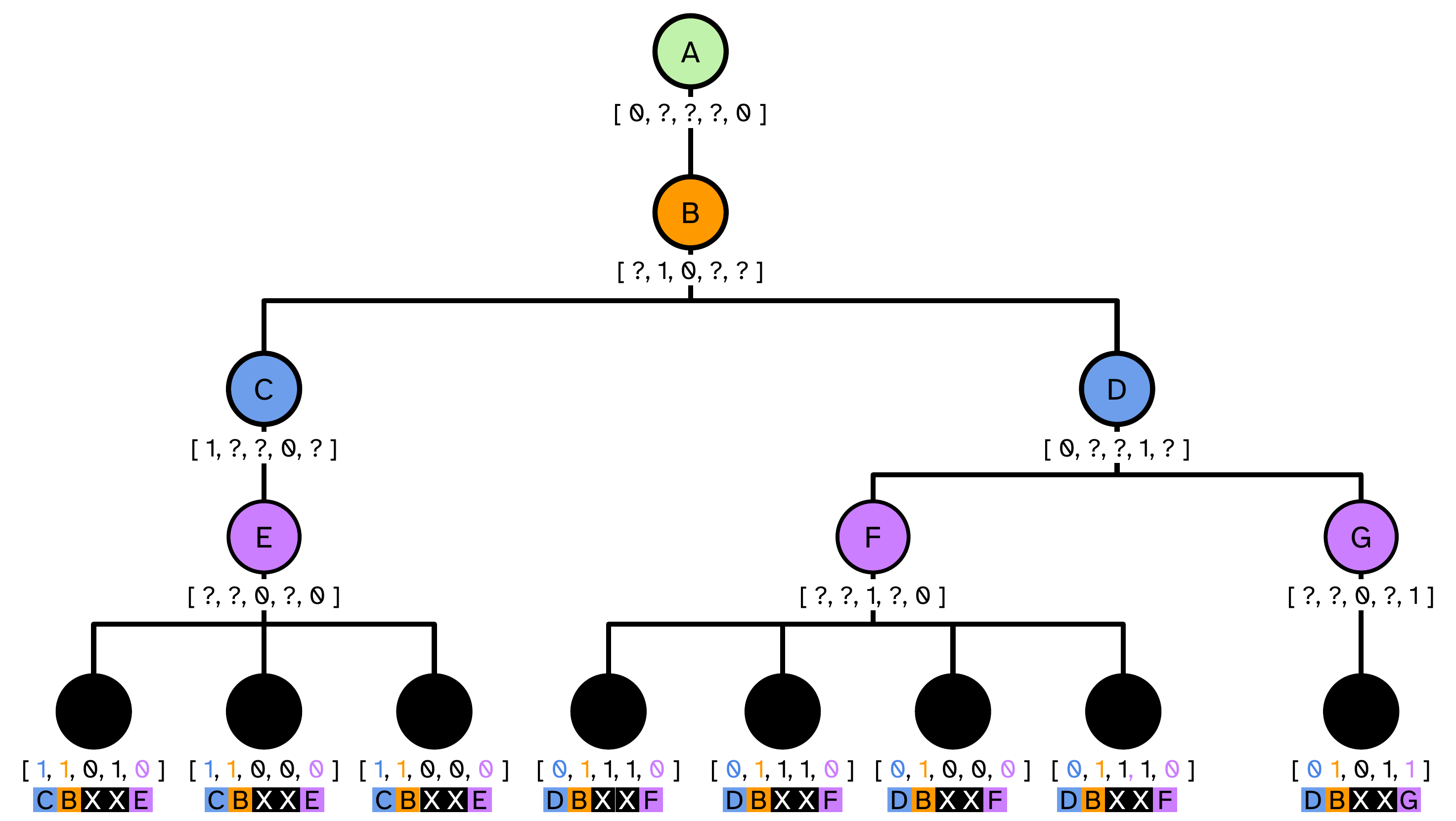}
\caption{\small{
Example application of ancestor-based phylogeny-informed fitness estimation. 
Taxa are depicted as circles in the tree. 
Extant taxa are unlabeled black circles (leaf nodes in the tree), and ancestral (non-extant) taxa are labeled A through G. 
In this example, an individual's quality is assessed using five pass/fail training cases. 
A taxon's scores on the five training cases are given as vectors below the associated node in the tree.
For ancestral taxa, a ``?'' indicates that the score was not evaluated (and is therefore unknown), a ``0'' indicates that taxon failed the training case when evaluted, and a ``1'' indicates that taxon passed the training case. 
In the extant population of this example, the third and fourth training cases were evaluated, and all others are estimated. 
For these extant taxa, ancestor-based estimates are shown in color (corresponding with the color of the ancestor used for the estimate), and evaluated scores are shown in black (with a ``X'' below).
}}
\label{fig:estimation-example}
\end{center}
\end{figure*}

%% file: tex/methods.tex
\section{Methods}
\label{sec:methods}

We investigated phylogeny-informed fitness estimation in the context of two variants of the lexicase parent selection algorithm that use random subsampling: down-sampled lexicase and cohort lexicase. 
For each of these selection schemes, we compared three different modes of fitness estimation: ancestor-based estimation, relative-based estimation, and a no-estimation control. 
We repeated all experiments at four different subsampling levels: 1\%, 5\%, 10\%, and 50\%.

First, we used selection scheme diagnostics from the DOSSIER suite~\cite{hernandez_suite_2022} to assess whether phylogeny-informed estimation is able to mitigate the drawbacks of random subsampling. 
These diagnostics measure different characteristics related to how a selection scheme steers populations through search spaces. 
Next, we investigated how each mode of phylogeny-informed estimation impacts problem-solving success on four genetic programming (GP) problems from the program synthesis benchmark suites \cite{helmuth_general_2015,helmuth_psb2_2021}.

\subsection{Lexicase selection}

The lexicase parent selection algorithm is designed for test-based problems where candidate solutions are assessed based on their performance on a set of input/output examples (training cases) that specify correct behavior~\cite{helmuth_solving_2015}.
To select a single parent, lexicase selection shuffles the set of training cases into a random order, and all members of the population are included in a pool of candidates eligible for selection. 
Each training case is then applied in sequence (in shuffled order), filtering the pool of eligible candidates to include only those candidates with elite performance on the current training case. 
This filtering continues until all training cases have been applied in the shuffled sequence. 
After filtering, if more than one candidate remains in the pool, one is selected at random. 

In practice, standard lexicase selection requires that all individuals in a population are evaluated against all training cases, which can be computationally expensive in many problem domains.
Down-sampled and cohort lexicase address this drawback by using random subsampling to reduce the required number of training case evaluations (and fitness comparisons) each generation. 

\subsubsection{Down-sampled lexicase selection}

Each generation, down-sampled lexicase selection randomly subsamples the training set, using only those sampled training cases for evaluation and selection that generation~\cite{hernandez_random_2019,moore_lexicase_2017}. 
When augmented with phylogeny-informed fitness estimation, down-sampled lexicase still evaluates the population on only the sampled training cases.
However, the full set of training cases is then used to select parents, and the population's phylogeny is used to estimate an individual's performance on any training cases it was not evaluated against.

\subsubsection{Cohort lexicase selection}

Cohort lexicase selection partitions the training set and the population into an equal number of ``cohorts''~\cite{hernandez_random_2019}.
We specify cohort sizes (subsample level) as proportions of the population and training set size, so a subsample level of 10\% would divide the population and training set into 10 evenly sized cohorts. 
Each generation, cohort membership is randomly assigned, and each cohort of candidate solutions is paired with a cohort of training cases. 
Then, each cohort of candidate solutions is evaluated only on the corresponding cohort of training cases. 
This method of partitioning reduces the number of per-generation evaluations required, but still uses the full set of training cases each generation.
To select a parent, cohort lexicase first selects a cohort to choose from, and applies standard lexicase selection to choose a parent from within the chosen cohort. 

When augmented with phylogeny-informed fitness estimation, cohort lexicase still evaluates the population as described above.
However, the full set of training cases is used when selecting a parent, and the population's phylogeny is used to estimate an individual's performance on any training cases it was not evaluated against. 

\subsection{Diagnostic experiments}

The DOSSIER suite comprises a set of diagnostic benchmarks used to empirically analyze selection schemes on important characteristics of evolutionary search~\cite{hernandez_suite_2022}.
This suite of diagnostics has already been used to illuminate differences between down-sampled and cohort lexicase, as well as a variety of other lexicase selection variants~\cite{hernandez_exploration_2022}, making it a good choice for analyzing the effects of phylogeny-informed fitness estimation. 
Here, we used the contradictory objectives and multi-path exploration diagnostics to isolate and measure the effect of phylogeny-informed fitness estimation on diversity maintenance and search space exploration. 
We parameterized these experiments similarly to the diagnostics experiments from~\cite{hernandez_suite_2022}. 

The diagnostics use a simple genome-based representation comprising a sequence of 100 floating-point values, each bound to the range 0.0 to 100.0. 
Each diagnostic specifies a translation function that translates a genome into a phenotype, which is also a length-100 sequence of numeric values.
Each position in a genome sequence is referred to as a ``gene'', and each position in a phenotype is a ``trait''. 
When using lexicase selection, each trait is treated as a training case where the trait's value is the candidate solution's score on the corresponding training case (higher scores are better). 
We apply subsampling to the diagnostics in the same way as in~\cite{hernandez_exploration_2022}; traits not included in a subsample are marked as ``unknown'', and their scores are not included in the phenotype. 

For all diagnostic experiments, we ran 10 replicates of each condition. 
In each replicate, we evolved a population of 500 individuals for 50,000 generations.
Each generation of a run, we evaluated the population on the appropriate diagnostic and used the condition-specific selection procedure to select 500 individuals to reproduce asexually. 
We mutated offspring at a per-gene rate of 0.7\%, drawing mutations from a normal distribution with a mean of 0.0 and a standard deviation of 1.0.  
For all conditions using phylogeny-informed fitness estimation, we limited phylogeny search depth to 10. 

Note that all subsampling levels are run for the same number of \textit{generations} in our diagnostic experiments, as we make comparisons across trait estimation methods (and not across subsample levels).   
We describe each of the three diagnostics below, but we direct readers to~\citep{hernandez_suite_2022} for more detailed descriptions. 

\subsubsection{Contradictory objectives diagnostic}

The contradictory objectives diagnostic measures how many global optima a selection scheme can find and maintain in a population. 
This diagnostic translates genomes into phenotypes by identifying the gene with the greatest value and marking that gene as ``active'' (and all others as ``inactive''). 
The active gene value is directly copied into the corresponding trait in the phenotype, and all traits associated with inactive genes are interpreted as zero-value traits. 
There are 100 independent global optima in this search space, one associated with each trait. 
All traits are maximized at the upper bound of 100.0. 
For analysis purposes, we label any trait above a score of 96 (out of 100) as ``satisfied''. 

On this diagnostic, we report ``satisfactory trait coverage''. 
Satisfactory trait coverage is a population-based measurement and gives the number of distinct traits labeled as satisfied across the entire population. 
Therefore, satisfactory trait coverage specifies the total number of unique global optima maintained in a population. %

\subsubsection{Multi-path exploration diagnostic}

The multi-path exploration diagnostic measures how well a selection scheme is able to continuously explore multiple pathways in a search space.
This diagnostic translates genomes into phenotypes by first identifying the gene with the greatest value and marking that genes as the ``activation gene''. 
Starting from the activation gene, all consecutive genes that are less than or equal to the previous gene are marked as active, creating an active region. 
Genes in the active region are directly copied into the corresponding traits in the phenotype, and all genes outside of the active region are interpreted as zero-valued traits. 
Traits are maximized at the upper bound (100.0), and a phenotype with all maximized traits occupies the global optimum in the search space. 

This diagnostic defines a search space with many pathways, each differing in path length and peak height but identical in slope (see Figure 5.1 in~\cite{hernandez_exploration_2022} for visual example). 
The pathway beginning at the first gene position leads to the global optimum, and all other pathways lead to local optima. 
Thus, a selection scheme must be able to continuously explore different pathways in order to consistently reach the global optimum. 
On this diagnostic, we report the best aggregate trait score in the final population; greater aggregate trait scores indicate that the population found better pathways in the search space.

\subsection{Genetic programming experiments}

We compared the problem-solving success of ancestor-based estimation, relative-based estimation, and a no estimation control on four GP problems from the first and second program synthesis benchmark suites~\cite{helmuth_general_2015,helmuth_psb2_2021}: Median, Small or Large, Grade, and Fizz Buzz.
As in our diagnostic experiments, we conducted our GP experiments using both down-sampled and cohort lexicase selection, each at four subsampling levels: 1\%, 5\%, 10\%, and 50\%.
For all conditions using phylogeny-informed fitness estimation, we limited phylogeny search depth to five. 

\subsubsection{GP system}

For all GP experiments, we ran 30 replicates of each condition.
In each replicate, we evolved a population of 1,000 linear genetic programs using the SignalGP representation~\cite{lalejini_evolving_2018}. 
The instruction set and complete configuration details (including source code) can be found in our supplemental material~\cite{supplemental_gh}.
SignalGP supports the evolution of programs composed of many callable modules; here, however, we limited all programs to a single program module for simplicity. 
We reproduced programs asexually and applied mutations to offspring.
Single-instruction insertions, deletions, and substitutions were applied, each at a per-instruction rate of 0.1\%.
We also applied `slip' mutations~\cite{lalejini_gene_2017}, which can duplicate or delete sequences of instructions, at a per-program rate of 5\%. 

\subsubsection{Program synthesis problems}

The first and second program synthesis benchmark suites include a variety of introductory programming problems that are well-studied and are commonly used to compare selection methods for GP. 
We chose the Median, Small or Large, Grade, and Fizz Buzz problems because they have training sets that contain different \textit{categories} of training cases.  
Brief descriptions of each problem are provided below: 

\begin{itemize}
    \item \textit{Median}~\cite{helmuth_general_2015}: 
    Programs are given three integer inputs ($-100 \le input_i \le 100$) and must output the median value. 
    We limited program length to a maximum of 64 instructions and limited the maximum number of instruction-execution steps to 64.  
    \item \textit{Small or Large}~\cite{helmuth_general_2015}: 
    Programs are given an integer $n$ and must output ``small'' if $n < 1000$, ``large'' if $n \ge 2000$, and ``neither'' if $1000 \ge n < 2000$.
    We limited program length to a maximum of 64 instructions and limited the maximum number of instruction-execution steps to 64. 
    \item \textit{Grade}~\cite{helmuth_general_2015}:
    Programs receive five integer inputs in the range [0, 100]: $A, B, C, D,$ and \textit{score}. 
    $A$, $B$, $C$, and $D$ are monotonically decreasing and unique, each defining the minimum score needed to receive that ``grade''.
    The program must read these thresholds and return the appropriate letter grade for the given \textit{score} or return $F$ if \textit{score} $< D$. 
    We limited program length to a maximum of 128 instructions and limited the maximum number of instruction-execution steps to 128.
    \item \textit{Fizz Buzz}~\cite{helmuth_psb2_2021}: 
    Given an integer $x$, the program must return ``Fizz'' if $x$ is divisible by 3, ``Buzz'' if $x$ is divisible by 5, ``FizzBuzz'' if $x$ is divisible by both 3 and 5, and $x$ if none of the prior conditions are true.
    We limited program length to a maximum of 128 instructions and limited the maximum number of instruction-execution steps to 128. 
\end{itemize}

For each problem, we used 100 training cases for program evaluation and selection and 1,000 testing cases used for determining problem-solving success; we provide the training and testing sets used for all problems in our supplemental material~\cite{supplemental_gh}.
We evaluated all training and test cases on a pass-fail basis. 
For all problems, we ensured that input-output edge cases were included in both the training and testing sets.
However, \textit{training sets} were not necessarily balanced; that is, we did not ensure equal representation of different categories of outputs. 
We categorized a replicate as successful if it produced a program that solved all training cases that it was evaluated against and all 1,000 testing cases. 
We terminated runs after a solution was found or after 30,000,000 training case evaluations, which represents 300 generations of evolution using \textit{standard} lexicase selection where all programs are evaluated against all training cases each generation. 

\subsection{Statistical analyses}

For all experiments, we did not compare measurements taken from treatments across problems, subsampling method (down-sampling and cohorts), or down-sampling level (1\%, 5\%, 10\%, and 50\%).
We made pairwise comparisons between fitness estimation methods that shared problem, subsampling method, and down-sampling level. 
When comparing distributions, we performed a Kruskal-Wallis test to look for statistical differences among independent conditions.
For comparisons in which the Kruskal-Wallis test was significant (significance level of 0.05), we performed a post-hoc Wilcoxon rank-sum test. 
When comparing problem-solving success rates, we used pairwise Fisher's exact tests (significance level of 0.05).
We used the Holm-Bonferroni method to correct for multiple comparisons where appropriate. 

\subsection{Software and data availability}

The software used to conduct our experiments, our data analyses, and our experiment data can be found in our supplemental material~\cite{supplemental_gh}, which is hosted on GitHub and archived on Zenodo. 
Our experiments are implemented in C++ using the Empirical library~\cite{ofria_empirical_2020}, and we used a combination of Python and R version 4 \cite{software_r} to conduct our analyses. 
Our experiment data are archived on the Open Science Framework at \url{https://osf.io/wxckn/}.

%% file: tex/results-and-discussion.tex
\section{Results and Discussion}
\label{sec:results-and-discussion}

\subsection{Phylogeny-informed estimation reduces diversity loss caused by subsampling}

Figure~\ref{fig:results-con-obj} shows the final satisfactory trait coverage achieved on the contradictory objectives diagnostic.  
The contradictory objectives diagnostic measures a selection scheme's ability to find and maintain many mutually exclusive optima.
Satisfactory trait coverage is the total number of unique global optima maintained in a population, which measures a selection scheme's capacity for diversity maintenance~\cite{hernandez_suite_2022}.
The horizontal line in Figure~\ref{fig:results-con-obj} indicates the median satisfactory trait coverage (of 10 runs) at the end of 50,000 generations of standard lexicase, representing a baseline reference for standard lexicase selection's capacity for diversity maintenance on this diagnostic. 

\input{tex/figures/fig-results-contradictory-objectives}

Both ancestor-based and relative-based estimation methods mitigated the diversity loss caused by subsampling across most subsampling levels for both down-sampled and cohort lexicase; though, the magnitude of mitigation varied by condition. 
Were were unable to detect a significant difference between ancestor-based estimation and the no estimation control for 1\% down-sampled lexicase, but in all other cases of down-sampled and cohort lexicase, both ancestor-based and relative-based estimation had greater satisfactory trait coverage (Corrected Wilcoxon rank-sum test, $p < 0.003$).
When applied to down-sampled lexicase at 50\% subsampling, phylogeny-informed estimation performed as well as our standard lexicase baseline, entirely mitigating the drawback of down-sampling.
Unsurprisingly, the efficacy of both phylogeny-informed estimation methods decreased as subsampling levels became more extreme. 

We initially expected relative-based estimation to facilitate better satisfactory trait coverage than ancestor-based estimation, as fitness estimates were not limited to direct ancestors. 
However, our data do not support this hypothesis. 
We were unable to detect a significant difference in satisfactory trait coverage between ancestor-based and relative-based estimation across all subsampling levels for both down-sampled and cohort lexicase.

Consistent with previous work~\cite{hernandez_exploration_2022}, cohort lexicase with no estimation generally maintained more objectives than down-sampled lexicase with no estimation. 
This is likely due to the fact that cohort lexicase uses \textit{all} training cases each generation (partitioned across different subsets of the population).
In contrast, down-sampled lexicase may omit an entire category of training cases for a generation.
Such omissions increase the chance that lexicase fails to select lineages that specialize on the excluded training cases, which can result in a loss of potentially important diversity.

Given that cohort lexicase uses all training cases every generation, we initially expected cohort lexicase to benefit from phylogeny-informed estimation more than down-sampled lexicase. 
Under cohort partitioning, nearby relatives in the phylogeny might be more likely to have been evaluated on a greater number of training cases, which could increase the accuracy of estimations. 
Observationally, however, cohort lexicase does not appear to have benefited from phylogeny-informed estimation more than down-sampled lexicase.  
Further investigation is needed to disentangle the interactions between subsampling methods and fitness estimation methods.

\subsection{Phylogeny-informed estimation improves poor exploration caused by down-sampling}

The multi-path exploration diagnostic measures a selection scheme's ability to explore many pathways in a search space in order to find the global optimum. 
Figure~\ref{fig:results-explore} shows performance (max aggregate score) results from all experiments with the multi-path exploration diagnostic. 
The horizontal line in Figure~\ref{fig:results-explore} indicates the median score (out of 10 runs) at the end of 50,000 generations of standard lexicase; this line provides a baseline reference for standard lexicase selection's capacity for search space exploration. 

\input{tex/figures/fig-results-exploration}

The phylogeny-informed estimation methods reduced the drawback of subsampling on the exploration diagnostic under most of the conditions tested, although the magnitude varied by condition. 
Under down-sampled lexicase, phylogeny-informed estimation improved performance at 5\%, 10\%, and 50\% subsampling levels (statistically significant, $p < 0.003$), and we were unable to detect a statistically significant difference between ancestor-based and relative estimation at any subsampling level. 
With phylogeny-informed estimation, some runs of 50\% down-sampled lexicase matched (or exceeded) the performance of our standard lexicase reference (Figure~\ref{fig:results-explore}a).
This is particularly notable, as we ran all conditions for an equal number of generations regardless of subsampling level; that is, down-sampling (at all subsampling levels) did not confer the usual benefits of increased population size or additional generations relative to our standard lexicase reference.

Under cohort lexicase, we were unable to detect statistically significant differences among conditions under 1\% and 50\% subsampling levels. 
Consistent with prior work~\cite{hernandez_exploration_2022}, the performance of cohort lexicase at moderate subsampling levels is close to that of standard lexicase on the exploration diagnostic, so there is little potential for either estimation method to improve over the no estimation control.
At 10\% subsampling (i.e., 10 partitions), ancestor-based and relative-based estimation outperformed the no estimation control ($p < 0.001$), and relative-based estimation outperformed ancestor-based estimation ($p < 0.001$). 
At 5\% cohort partitioning (i.e., 20 partitions), relative-based estimation outperformed both ancestor-based estimation and the no estimation control ($p < 0.05$), but we detected no statistically significant difference between ancestor-based estimation and the no estimation control. 

As noted previously, we held \textit{generations} constant across subsampling levels for these experiments.
We do expect performance to continue to increase given longer runtimes  (performance over time is visualized in the supplemental material~\cite{supplemental_gh}).
Further experimentation is needed to investigate whether extra generations would allow for smaller subsample sizes (e.g., 1\%, 5\%) to achieve the performance of more moderate subsample sizes or of full lexicase selection. 
Moreover, it is unclear whether phylogeny-informed estimation would allow for more rapid progress over no estimation controls if given more generations of evolution. 
 
Phylogeny-informed estimation more consistently improved performance under down-sampled lexicase (3 out of 4 subsampling levels) than under cohort lexicase (2 out of 4 subsampling levels). 
We hypothesize that cohort partitioning may interact poorly with our implementation of phylogeny-informed estimation in the context of lexicase selection.
Lexicase selection is sensitive to the ratio between population size and the number of training cases used for selection \cite{hernandez_exploration_2022}.
Increasing the number of training cases for a fixed population size decreases the odds that any particular training case will appear early in any of the shuffled sequences of training cases used during selection, which can be detrimental to lexicase's ability to maintain diversity.
Cohort partitioning reduces the effective population size for any given selection event, as candidate solutions compete within only their cohort.
Normally, cohort partitioning also reduces the number of training cases used to mediate selection events within each cohort, maintaining approximately the same ratio between population size and the number of training cases used for selection.
However, in our implementation of phylogeny-informed estimation, we use \textit{all} training cases to mediate selection in each cohort; during selection, this changes the ratio between the population size and number of training cases used for selection. 
We hypothesize that this dynamic can cause phylogeny-informed estimation to underperform when used in combination with cohort lexicase selection, especially for problems with low levels of redundancy in the training cases. 
To mitigate this problem, we could modify cohort partitioning when used in combination with phylogeny-informed estimation: evaluation can be conducted according to standard cohort partitioning procedures (still using the full training set across the entire population), but when selecting parents, we can allow all individuals in the population to compete with one another (as opposed to competing only within their cohort).

\subsection{Phylogeny-informed estimation can enable extreme subsampling for some genetic programming problems}

Tables~\ref{tab:results-psynth}a and~\ref{tab:results-psynth}b show problem-solving success on the Median, Small or Large, Grade, and Fizz Buzz program synthesis problems for down-sampled lexicase and cohort lexicase, respectively.
We consider a run to be successful if it produces a program capable of passing all tests in the testing set.
In total, we compared the success rates of ancestor-based estimation, relative-based estimation, and a no estimation control under 32 distinct combinations of subsampling method (down-sampling and cohort partitioning), subsampling level (1\%, 5\%, 10\%, and 50\%), and program synthesis problem.

\input{tex/figures/tab-results-psynth}

We were unable to detect a statistically significant difference among fitness estimation treatments in 20 out of 32 comparisons (Table~\ref{tab:results-psynth}). 
At least one phylogeny-informed estimation method resulted in greater success rates in 4 out of 16 comparisons under down-sampled lexicase and in 5 out of 16 comparisons under cohort lexicase ($p < 0.05$, Fisher's exact test with Holm-Bonferroni correction for multiple comparisons). 
Under down-sampled lexicase, the no estimation control outperformed at least one phylogeny-informed estimation method in 3 out of 16 comparisons ($p < 0.04$).
Under cohort lexicase, we detected no instances where the no estimation control resulted in a significantly greater number of successes than either phylogeny-informed estimation method. 

Overall, the benefits of phylogeny-informed estimation varied by subsampling method and by problem.
Additionally, we found no statistically significant differences in success rates between ancestor-based and relative-based estimation across all comparisons. 
Under cohort lexicase, phylogeny-informed estimation had a more consistently neutral or beneficial effect on problem-solving success. 
Down-sampled lexicase, however, was sometimes more successful without the addition of phylogeny-informed estimation; for example, on the Fizz Buzz problem with 5\% down-sampling, the no estimation control substantially outperformed either phylogeny-informed estimation control (Table~\ref{tab:results-psynth}a).
This result contrasts with the diagnostics results where phylogeny-informed estimation was more generally helpful. 
We expect that this difference is rooted in the fact that we looked at a limited set of diagnostics, each of which isolate specific problem-solving characteristics.  

Our diagnostics show that phylogeny-informed estimation can improve diversity maintenance and multi-path exploration when subsampling; however, there are other problem-solving characteristics that our initial methods of phylogeny-informed estimation may degrade. 
For example, any fitness estimation technique can result in inaccurate estimates, which may result in the selection of low quality candidate solutions that would not have been otherwise selected.
Further investigation is required to disentangle how phylogeny-informed estimation can impede evolutionary search, which will allow us to improve our estimation methods to address these shortcomings.
Future work will run phylogeny-informed estimation on the full suite of DOSSIER diagnostic problems, assessing a wider range of problem-solving characteristics.
Additionally, further analysis of genetic programming conditions where phylogeny-informed estimation performs poorly may reveal additional issues.   

Problem-solving success at 1\% subsampling is particularly impressive, as evaluation was limited to just one training case per individual. 
Under down-sampled lexicase, 1\% subsampling results in elite selection mediated by the sampled training case, and under cohort lexicase, each of the 100 cohorts undergo elite selection on the single training case associated with each. 
Even under these seemingly extreme conditions, the no estimation control produced solutions to the median and grade problems.
Phylogeny-informed estimation methods more consistently improved problem-solving success at this extreme subsampling level, finding at least one solution to each of the four problems (across subsampling methods).

%% file: tex/figures/fig-results-contradictory-objectives.tex
\begin{figure*}[htb!]
\begin{center}
\includegraphics[width=\textwidth]{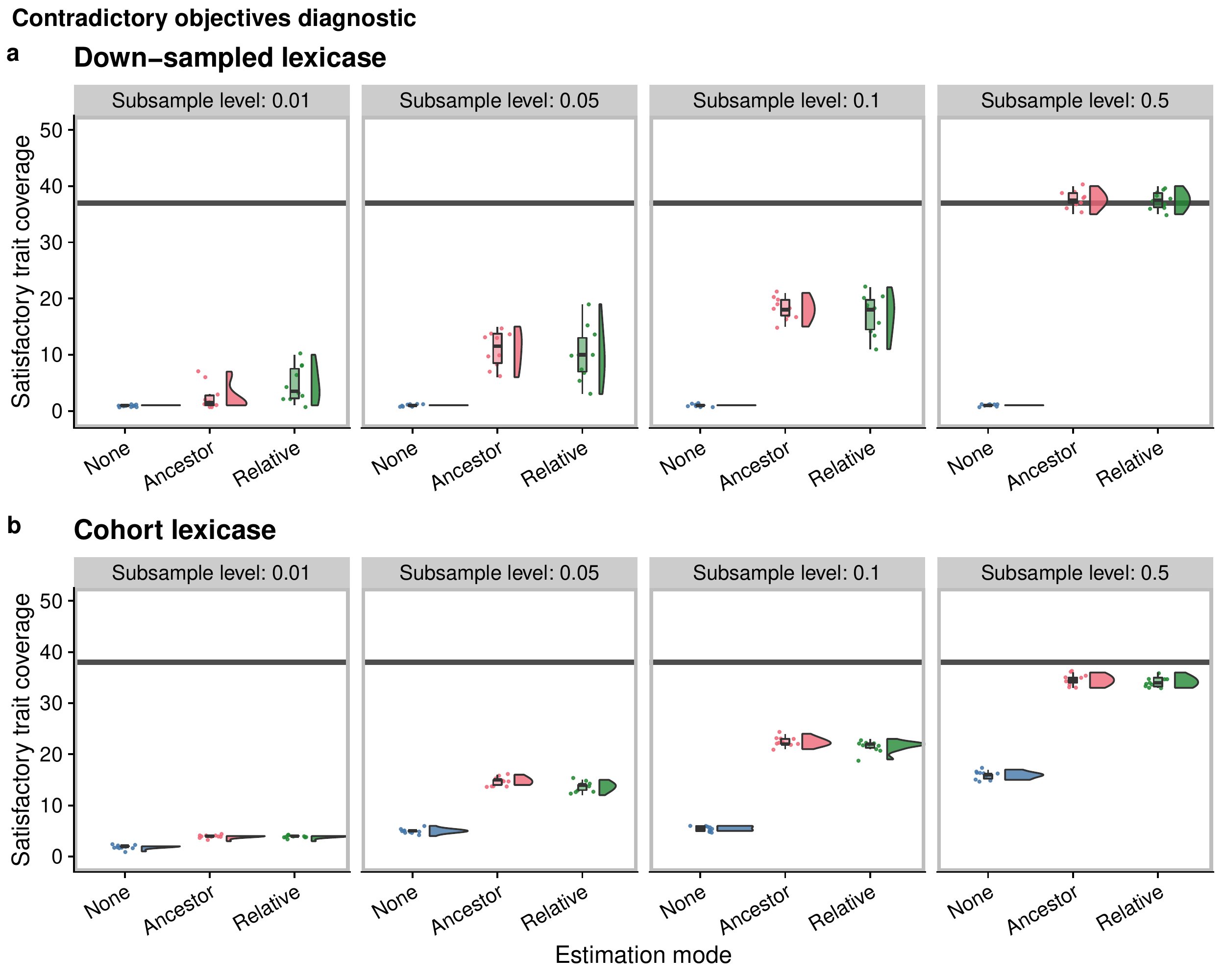}
\captionsetup{belowskip=0pt}
\caption{\small{
Satisfactory trait coverage on the contradictory objectives diagnostic. 
Panels (a) and (b) show satisfactory trait coverage for ancestor-based, relative-based, and no estimation in the context of down-sampled lexicase and cohort lexicase, respectively. 
The black horizontal line in each plot indicates the median satisfactory trait coverage of 10 runs of standard lexicase selection (no subsampling).
Kruskal-Wallis tests for all subampling levels for each of down-sampled and cohort lexicase were statistically significant ($p < 0.001$).
}}
\label{fig:results-con-obj}
\end{center}
\end{figure*}

%% file: tex/figures/fig-results-exploration.tex
\begin{figure*}[hbt!]
\begin{center}
\includegraphics[width=\textwidth]{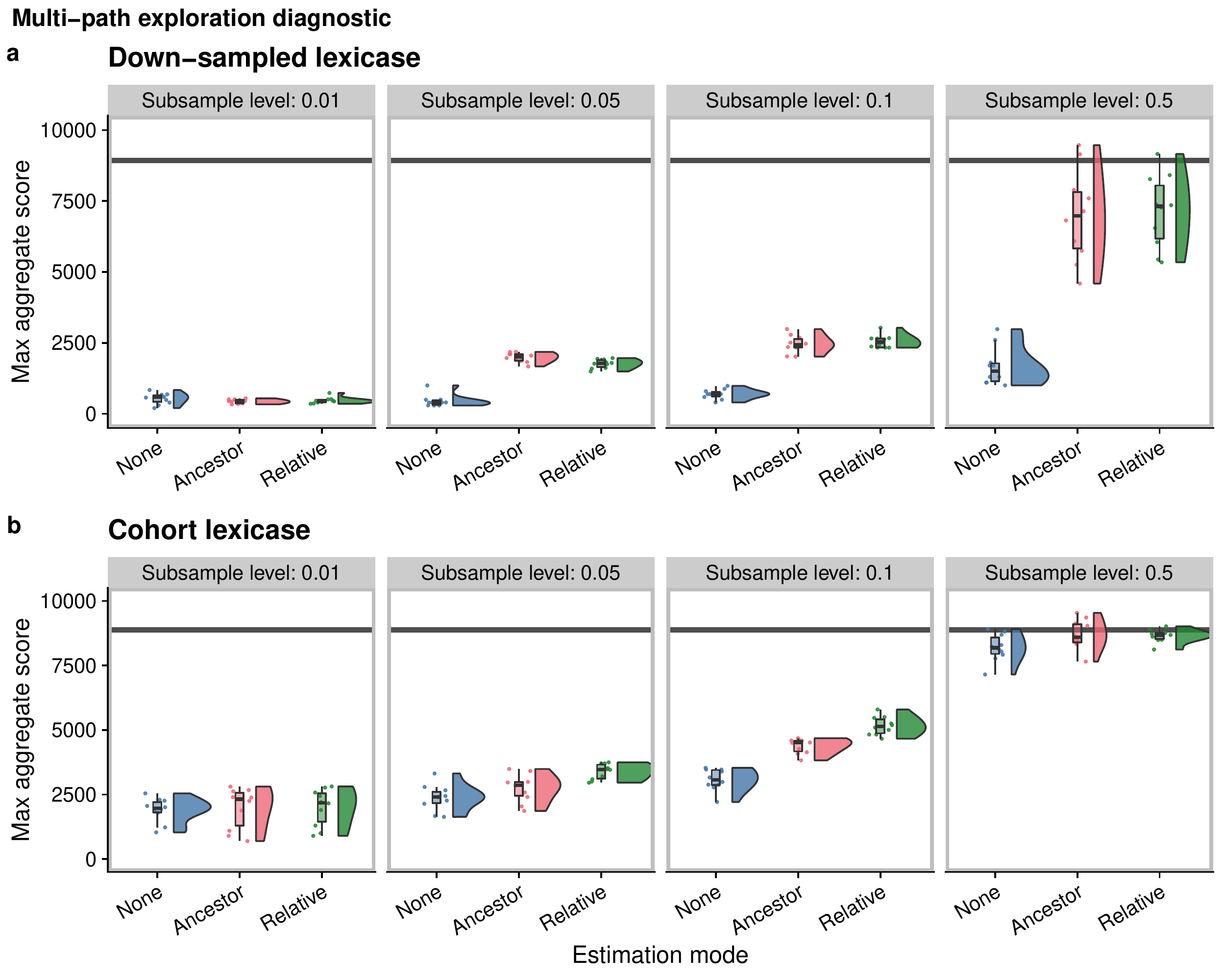}
\caption{\small{
Maximum aggregate scores in the final generation on the multi-path exploration diagnostic. 
Panels (a) and (b) show max aggregate scores for ancestor-based, relative-based, and no estimation in the context of down-sampled lexicase and cohort lexicase, respectively. 
The black horizontal line in each plot indicates the median max aggregate score from 10 runs of standard lexicase selection (no subsampling).
Kruskal-Wallis tests for each of the following comparisons were statistically significant ($p < 0.001$): 
down-sampled lexicase at 5\%, 10\%, and 50\% subsampling levels
and
cohort lexicase at 5\% and 10\% subsampling levels.
}}
\label{fig:results-explore}
\end{center}
\end{figure*}

%% file: tex/figures/tab-results-psynth.tex
\begin{table}[ht]
    \centering
    \caption{\small{
    Problem-solving success for down-sampled lexicase (a) and cohort lexicase (b) on GP problems, Median, Small or Large, Grade, and Fizz Buzz. 
    For each subsampling level, we compared the success rates of ancestor-based estimation, relative-based estimation, and no estimation (out of 30 replicates each).
    The ``Full'' column gives the success rate of running standard lexicase selection (out of 30 replicates); we include these results as a baseline reference and did not make direct comparisons to them. 
    We were unable to detect a statistically significant difference between relative-based and ancestor-based estimation across all tables (Fisher's exact test with a Holm-Bonferroni correction for multiple comparisons). 
    If phylogeny-based estimation results are bolded and italicized, they each significantly outperformed the no estimation control. 
    If a no estimation control result is bolded and italicized, it significantly outperformed both phylogeny-based estimation methods. 
    A result is annotated with a \textdagger\ if that condition performed significantly better than one other condition but not both other conditions. 
    }}
    \label{tab:results-psynth}
    \includegraphics[width=0.9\textwidth]{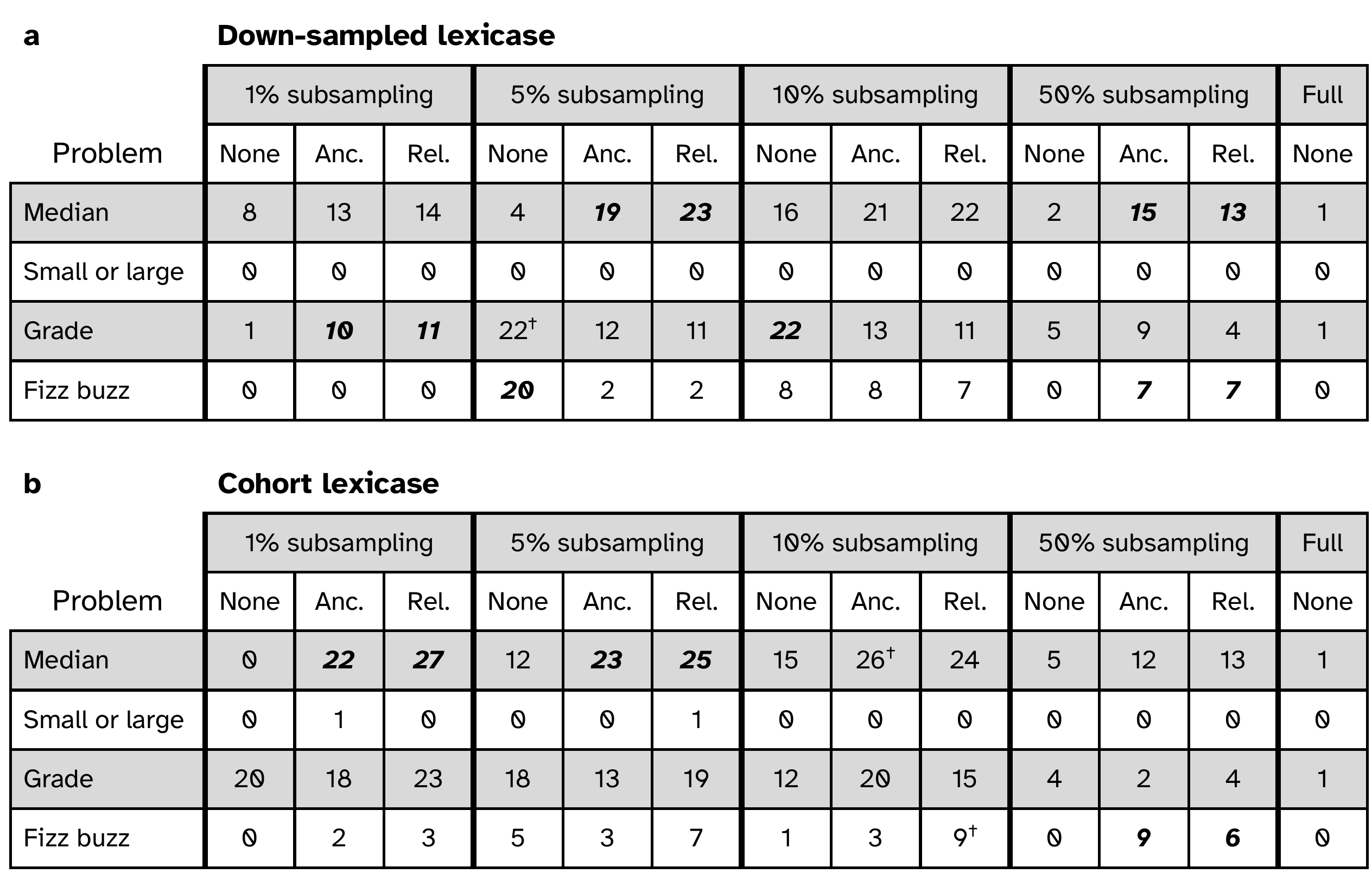}
\end{table}

%% file: tex/conclusion.tex
\section{Conclusion}
\label{sec:conclusion}

In this work, we take an initial step in exploiting runtime phylogeny tracking to help steer an evolutionary search. 
Specifically, we demonstrate two preliminary methods of phylogeny-informed fitness estimation in the context of cohort lexicase selection and down-sampled lexicase selection: ancestor-based estimation and relative-based estimation.
Using selection scheme diagnostic problems, we find evidence that our phylogeny-informed estimation methods can help to mitigate the drawbacks of random subsampling in the context of lexicase selection, improving diversity maintenance (Figure~\ref{fig:results-con-obj}) and search space exploration (Figure~\ref{fig:results-explore}).
We also find that phylogeny-informed estimation's effects on problem-solving success for GP varies by problem, subsampling method, and subsampling level (Table~\ref{tab:results-psynth}).

Overall, we did not find consistent or substantial differences between ancestor-based and relative-based fitness estimation. 
Future work will continue to disentangle how these two estimation methods influence evolutionary search. 
As of now, we recommend the use of ancestor-based estimation, as its implementation can be more effectively optimized than relative-based estimation. 

We opted for simplicity in our application of phylogeny-informed estimation to both down-sampled lexicase and cohort lexicase. 
Estimation methods allow each lexicase parent selection event to use the full training set, as an individual's performance on any unevaluated training cases can be estimated.
As such, phylogeny-informed estimation allows each member of the population to be evaluated on a different subset of the training set. 
We could subsample the training cases to minimize the distance in the phylogeny required for estimates, which may lead to an increase in estimation accuracy and improved problem-solving success.
Building on this idea, we could also take into account mutation information when choosing subsamples.
For example, we could more thoroughly evaluate (i.e., use more training cases) offspring with larger numbers of mutations, as they may be more likely to be phenotypically distinct from their parents. 

In addition to the phylogeny-informed estimation methods proposed in this work, we envision many ways in which runtime phylogeny tracking can be used to improve evolutionary search.
Here, we augmented existing subsampling methods, but we could design new subsampling methods optimized for use with phylogeny-informed fitness estimation.
For example, we could focus subsampling on training cases that an individual's ancestors have not been recently evaluated against. 
Beyond fitness estimation, runtime phylogenetic analyses may be useful in the context of many quality diversity algorithms, which currently rely on phenotypic or behavioral diversity.

%% file: tex/acknowledgements.tex
\section*{Acknowledgements}

We thank the participants of Genetic Programming Theory and Practice XX for helpful comments and suggestions on our work. 
In particular, we thank Nicholas McPhee for insightful feedback on our manuscript.
This work was supported in part through computational resources and services provided by the Institute for Cyber-Enabled Research at Michigan State University.